\documentclass[lettersize,journal]{IEEEtran}
\usepackage{amsmath,amsfonts}
\usepackage{algorithm}
\usepackage{algpseudocode}
\usepackage{array}
\usepackage[caption=false,font=normalsize,labelfont=sf,textfont=sf]{subfig}
\usepackage{textcomp}
\usepackage{stfloats}
\usepackage{url}
\usepackage{verbatim}
\usepackage{graphicx}
\usepackage{cite}
\usepackage{xcolor}
\usepackage[capitalise]{cleveref}
\begin{document}

\title{
Learning Probabilistic Obstacle Spaces from Data-driven Uncertainty using Neural Networks}

\author{Jun Xiang,
        Jun~Chen,~\IEEEmembership{Senior Member,~IEEE}
\thanks{The authors are with the Department of Aerospace Engineering, San Diego State University, San Diego, CA 92182 USA. (emails: jxiang9143@sdsu.edu; jun.chen@sdsu.edu).}
}


    
\maketitle

\begin{abstract}
Identifying the obstacle space is crucial for path planning. However, generating an accurate obstacle space remains a significant challenge due to various sources of uncertainty, including motion, behavior, and perception limitations. 
Even though an autonomous system can operate with an inaccurate obstacle space by being over-conservative and using redundant sensors, a more accurate obstacle space generator can reduce both path planning costs and hardware costs. Existing generation methods that generate high-quality output are all computationally expensive. Traditional methods, such as filtering, sensor fusion and data-driven estimators, face significant computational challenges or require large amounts of data, which limits their applicability in realistic scenarios. In this paper, we propose leveraging neural networks, commonly used in imitation learning, to mimic expert methods for modeling uncertainty and generating confidence regions for obstacle positions, which we refer to as the probabilistic obstacle space. The network is trained using a multi-label, supervised learning approach. We adopt a fine-tuned convex approximation method as the expert to construct training datasets. After training, given only a small number of samples, the neural network can accurately replicate the probabilistic obstacle space while achieving substantially faster generation speed. Moreover, the resulting obstacle space is convex, making it more convenient for subsequent path planning. 
\end{abstract}

\begin{IEEEkeywords}
Path planning, obstacle space, uncertainty, neural networks
\end{IEEEkeywords}

\section{Introduction}
\IEEEPARstart{A}{utonomous} vehicles operations face various uncertainties and hazards~\cite{thompson2022survey}. One of the most important tasks of any autonomous vehicle is to move without collisions, in other words, to remain within a safe space at all times~\cite{dawson2020provably}. 
Safe space is also called safe region~\cite{7784290}, obstacle-free space~\cite{wang2025fast} in other works. The first step to find a safe space is usually to find the obstacle space. 
While many methods exist for generating a safe space under the assumption of a known obstacle space~\cite{savic2015target}, this assumption is often unrealistic in practice, and only a few approaches~\cite{axelrod2018provably, shimanuki2021hardness} extend to uncertain obstacles. Meanwhile, many previous path planning algorithms assume the obstacle space is polygonal~\cite{deits2015computing, marcucci2023motion} due to the simplicity of collision checking and the compactness of the representation. The area of the obstacle space should be minimized, as a larger safe space typically results in smoother and shorter trajectories while also reducing the overall planning cost~\cite{wang2025fast}. Building on these insights, this paper proposes a novel method to construct a polygonal \textit{probabilistic obstacle space} that is both compact and capable of representing uncertain obstacles.


In practice, safety is ensured by declaring a collision whenever the distance between an agent and an obstacle falls below a threshold, even without physical contact~\cite{7857061}. In this work, we focus on obstacles with relatively small volumes, such as flights or UAVs in airspace~\cite{8385188}, where the obstacle space is determined primarily by position rather than precise geometry, a common simplification known as the point mass model. An obstacle is certain if a deterministic value represents its position, and uncertain if a probability distribution describes its position. The probabilistic obstacle space is then defined as the confidence region of this distribution.

Uncertainty of an obstacle can be caused by many reasons. The first common reason is perception uncertainty. All of the position-locating sensors, including GPS~\cite{ferguson2000global}, camera~\cite{zhu2020quantifying}, Lidar~\cite{hassani2021new}, and radar~\cite{hong2023large}, are subject to measurement noise. It is necessary to construct a probabilistic obstacle space from noisy measurements. Sensor Uncertainty Mitigation (SUM) methods~\cite{abramson2020applying} can compute bounds on the errors, but it has a very high computational cost and require diverse and representative data. Meanwhile, those over-conservative approaches make safe spaces very small, making it difficult for planning algorithms to find feasible solutions



Behavioral uncertainty can also render an obstacle uncertain, since different intents lead to entirely different trajectories~\cite{10160273}. Traditional data-driven methods~\cite{xiang2025data, zhi2021probabilistic} can estimate intent distributions when sufficient historical data are available. However, such distributions are often highly complex and require many parameters to represent,  which makes it impractical to directly translate them into obstacle spaces.


Another common source of uncertainty is motion. Even obstacles following a fixed path can exhibit positional variability~\cite{axelrod2018provably, miura2000modeling}. For instance, as shown in Fig.~\ref{fig:takeoff}, a flight may follow its planned trajectory yet still wander around it. Consequently, even with state-of-the-art trajectory predictors, the error between the predicted and actual positions can never be eliminated. The M-estimator of Tukey~\cite{1360125} has been used to estimate obstacle motion from single-camera images, but it is computationally expensive and requires careful parameter tuning.



In this paper, we propose a supervised learning approach that employs multi-label training to train a transformer-based neural network for generating polygonal probabilistic obstacle spaces under uncertainty. Numerical experiments demonstrate that the network can reliably reproduce probabilistic obstacle spaces when the input samples are drawn from obstacle types it has learned. In addition, the proposed method achieves substantially faster generation times compared to baseline approaches. The method offers four key advantages:

\begin{itemize}
\item \textit{Advantage 1}: The approach does not rely on dynamic models of uncertain obstacles. Instead, it can construct obstacle spaces from only a few noisy measurements or partial observations, reducing the sensing and modeling requirements.
\item \textit{Advantage 2}: All generated obstacle spaces are convex and linear, which simplifies their integration into path planning algorithms and ensures computational tractability.
\item \textit{Advantage 3}: Training requires only a small amount of labeled data, which lowers the burden of data collection and annotation and makes the approach practical for large-scale deployment.
\item \textit{Advantage 4}: The generation process is extremely fast, enabling real-time use in online planning and decision-making systems.
\end{itemize}



\section{Problem Statement}

The main goal of this paper is to generate the \textit{probabilistic obstacle space} of an uncertain obstacle from a \textit{limited data sample} using a \textit{neural network}:

\begin{equation}\label{eq: nn}
o^{\alpha} = NN(w, \mathbf{s}),
\end{equation}

where $o^{\alpha}$ is the probabilistic obstacle space at confidence level $\alpha \in (0,1)$, $w$ denotes the neural network parameters, and $\mathbf{s} = {s^i}_{i=1}^n$ is the set of observed $n$ samples.

\textbf{Definition (Probabilistic Obstacle Space).}
Let $o^{true}$ be the true occupied region of an uncertain obstacle $o$, and let $\mathbb{P}$ denote the underlying probability measure that characterizes uncertainty in the obstacle’s position. A set $o^{\alpha}$ is called a probabilistic obstacle space at confidence level $\alpha$ if it satisfies

\begin{equation}\label{eq: prob_def}
\mathbb{P}[a \in o^{true} \mid a \notin o^{\alpha}] < 1 - \alpha.
\end{equation}

That is, if the agent $a$ avoids $o^{\alpha}$, then the probability of entering the true obstacle region $o^{true}$ is bounded above by $1 - \alpha$.

Since $o^{true}$ is generally unknown due to imperfect perception and prediction, $o^{\alpha}$ provides a tractable confidence-region approximation. Our definition generalizes several prior notions, including the $\epsilon$-shadow of obstacle spaces~\cite{axelrod2018provably, shimanuki2021hardness}, Gaussian-distributed obstacles~\cite{wu2020probabilistically}, adaptive bounding boxes~\cite{timans2024adaptive}, and $\alpha$-probabilistic occupied sets~\cite{9157911}.

The neural network in Eq.~\eqref{eq: nn} is thus trained to approximate the mapping
\begin{equation}
f: \mathcal{S}^n \to \mathcal{O}, \quad f(\mathbf{s}) \mapsto o^{\alpha},
\end{equation}
where $\mathcal{S}^n$ is the sample space of $n$ measurements and $\mathcal{O}$ is the family of convex polygonal sets in $\mathbb{R}^d$.

We illustrate three representative examples of probabilistic obstacle spaces and their associated data samples.

\begin{figure}
    \centering
    \includegraphics[width=0.95\linewidth]{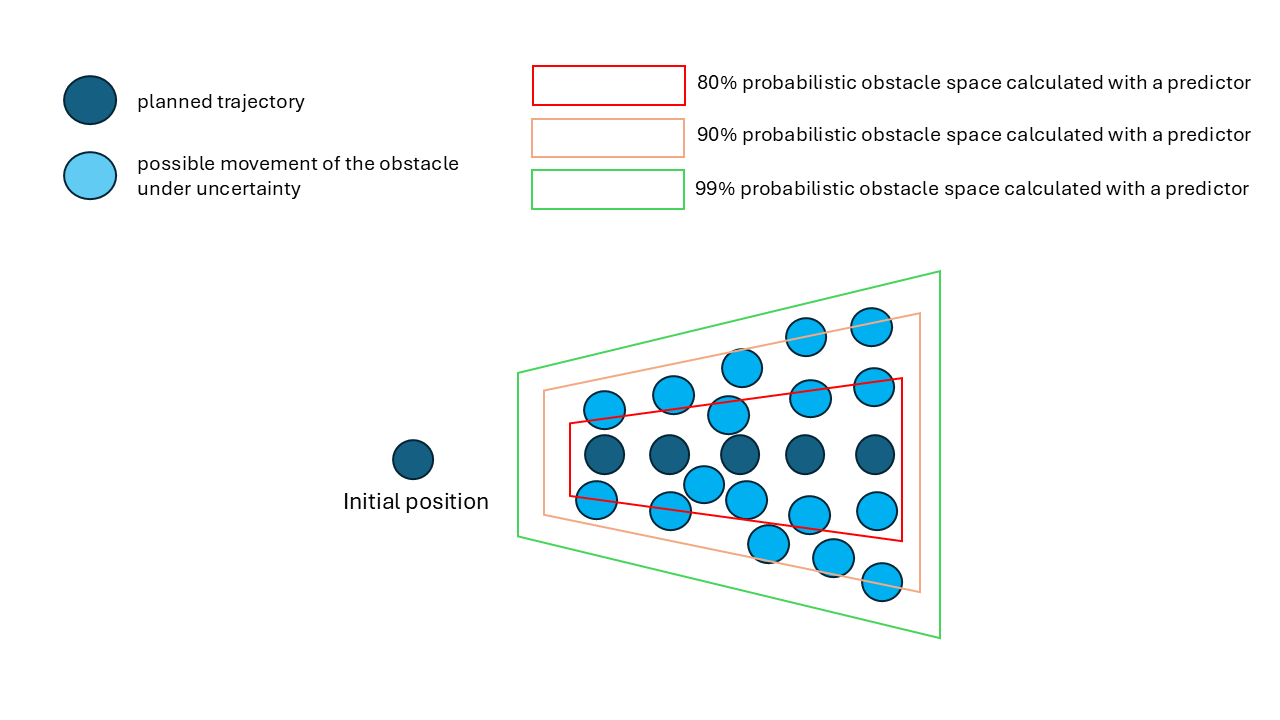}
    \caption{Motion uncertainty: even with a planned trajectory, the actual obstacle position is affected by disturbances and uncertainties.}
    \label{fig:followpath}
\end{figure}
Motion uncertainty: Consider the scenario of a moving object following a fixed planned trajectory. However, many factors, such as environmental disturbances, control uncertainties, and human factors, can introduce uncertainty in the actual trajectory. Therefore, as shown in Fig.~\ref{fig:followpath}, a single planned trajectory may correspond to infinitely many possible realizations of motion, all of which the ego agent must avoid. 
To capture this variability, we construct a probabilistic obstacle space that encompasses the possible deviations. Formally, we define the sample set $\mathbf{s} = {s^i}_{i=1}^n$ as the trajectories of $n$ flights following the same planned path.

\begin{figure}
    \centering
    \includegraphics[width=0.95\linewidth]{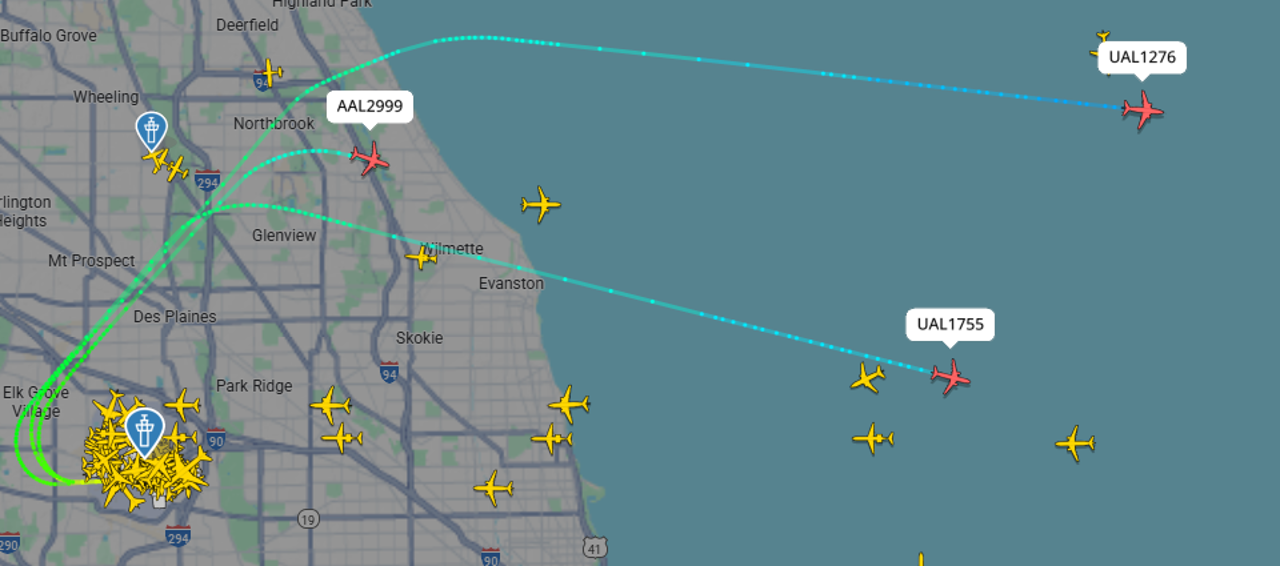}
    \includegraphics[width=0.95\linewidth]{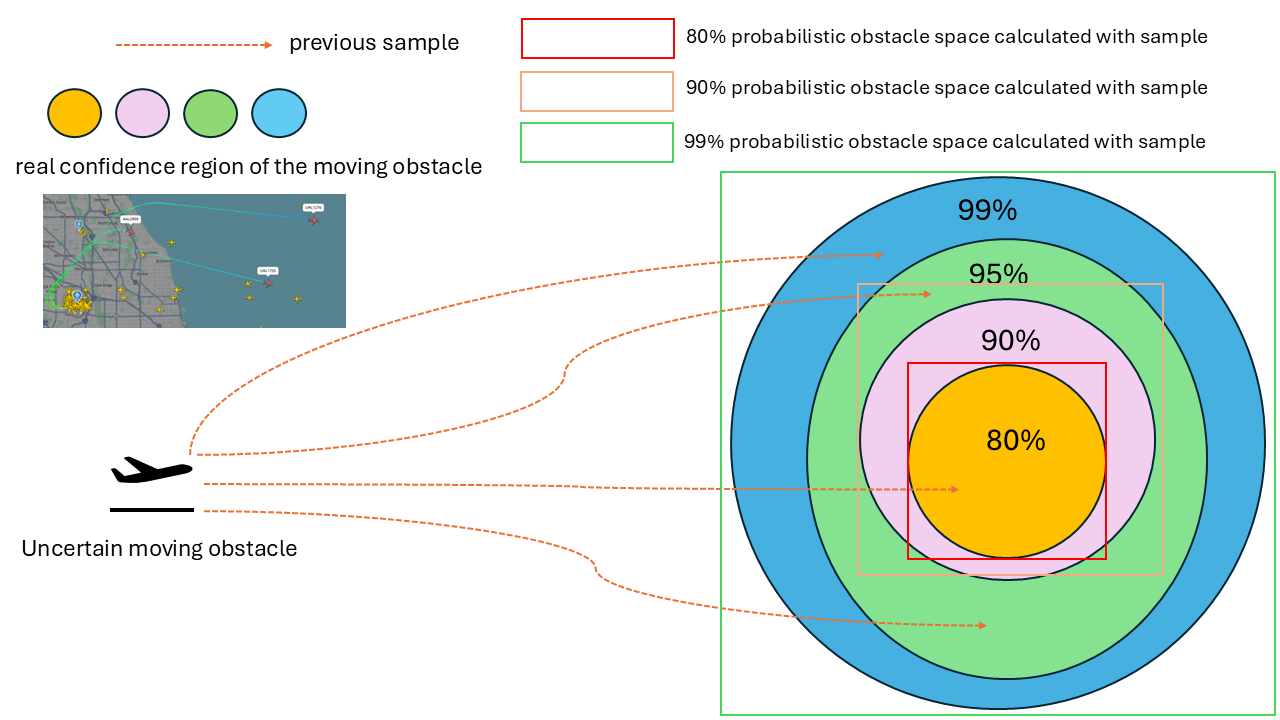}
    \caption{Behavior uncertainty: flights taking off from the same runway with a different destination will have a different route}
    \label{fig:takeoff}
\end{figure}

Behavior uncertainty: Another important source of uncertainty arises from the intent of other agents. Consider an ego UAV operating in terminal airspace near a busy airport. Although aircraft follow air traffic control clearances and standard procedures, the UAV does not know in advance which clearance each departure will receive. As shown in Fig.~\ref{fig:takeoff}, multiple aircraft may depart from the same runway within a narrow time window, and each departure may be cleared for a different standard instrument departure depending on destination, weather, and traffic flow. From the UAV’s perspective, the resulting obstacle space is therefore governed by a probability distribution over several candidate routes rather than a single deterministic path. Because such intent distributions are complex and rarely available in closed form, we approximate them by constructing the probabilistic obstacle space from $\mathbf{s}$, defined as the trajectories of $n$ flights that have previously departed from the same runway.

\begin{figure}
    \centering
    \includegraphics[width=0.95\linewidth]{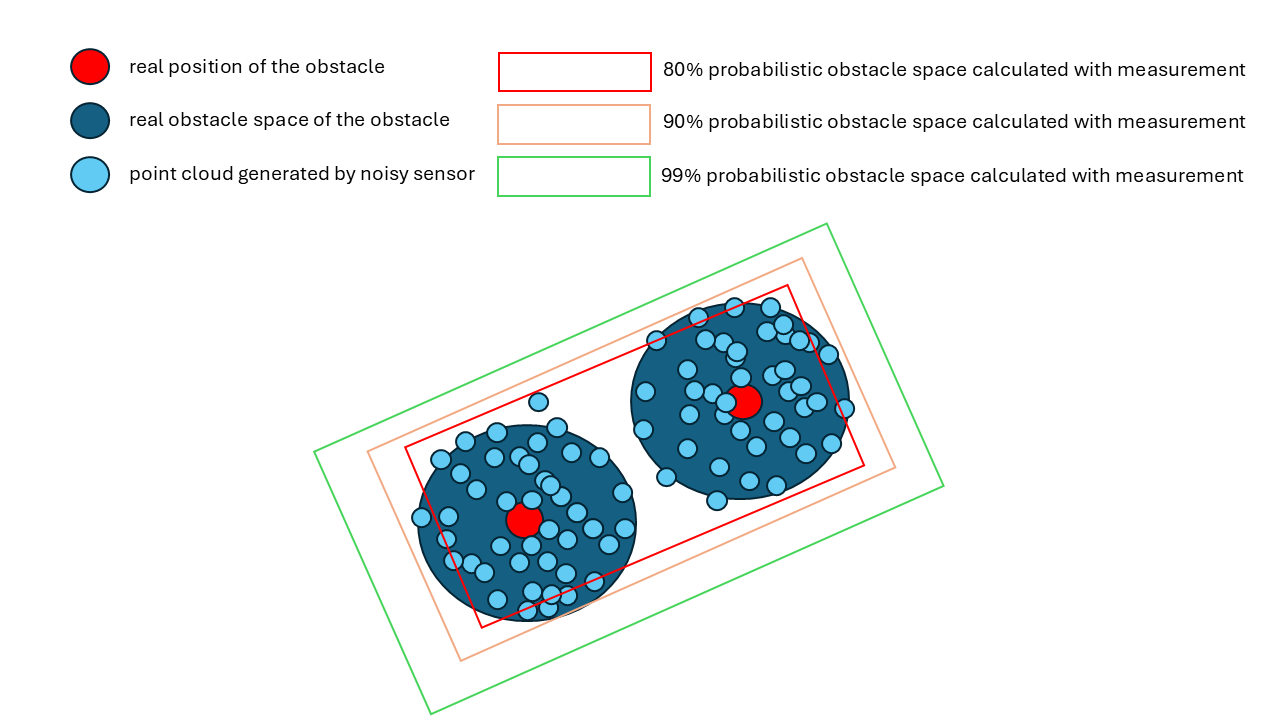}
    \caption{Sensor uncertainty: the position of two adjacent obstacles whose positions are estimated using noisy sensor measurements is uncertain}
    \label{fig:noisypointcloud}
\end{figure}

Sensor uncertainty: 
Fig.~\ref{fig:noisypointcloud} illustrates two adjacent circular obstacles whose positions are estimated using noisy sensor data. Because the measurements are corrupted by noise, the exact obstacle positions cannot be determined without calibration or advanced signal processing techniques. Furthermore, a single measurement outcome may be consistent with multiple possible true obstacle positions. To account for this, we define $\mathbf{s}$ as the set of $n$ noisy sensor measurements, from which a probabilistic obstacle space is constructed.


Since the obstacle space is inherently probabilistic, its exact boundary may be irregular or complex. For path planning, however, a tractable geometric representation is preferred. In particular, polygonal approximations are widely used because they provide a convex, linear structure that can be efficiently integrated into optimization-based planners. To this end, instead of directly generating the full probabilistic obstacle space, we represent it using a convex polygon defined by its vertices. Specifically, the neural network outputs four vertices that characterize the polygonal approximation of the probabilistic obstacle space:

\begin{equation}\label{eq: polygon}
\tilde{\mathcal{V}} = NN(w, \mathbf{s}),
\end{equation}

where $w$ denotes the trained parameters of the neural network and $\tilde{\mathcal{V}}$ is the predicted set of polygon vertices.

\section{Method}
\subsection{Supervised learning and expert method}
\begin{figure*}
    \centering
    \includegraphics[width=\linewidth]{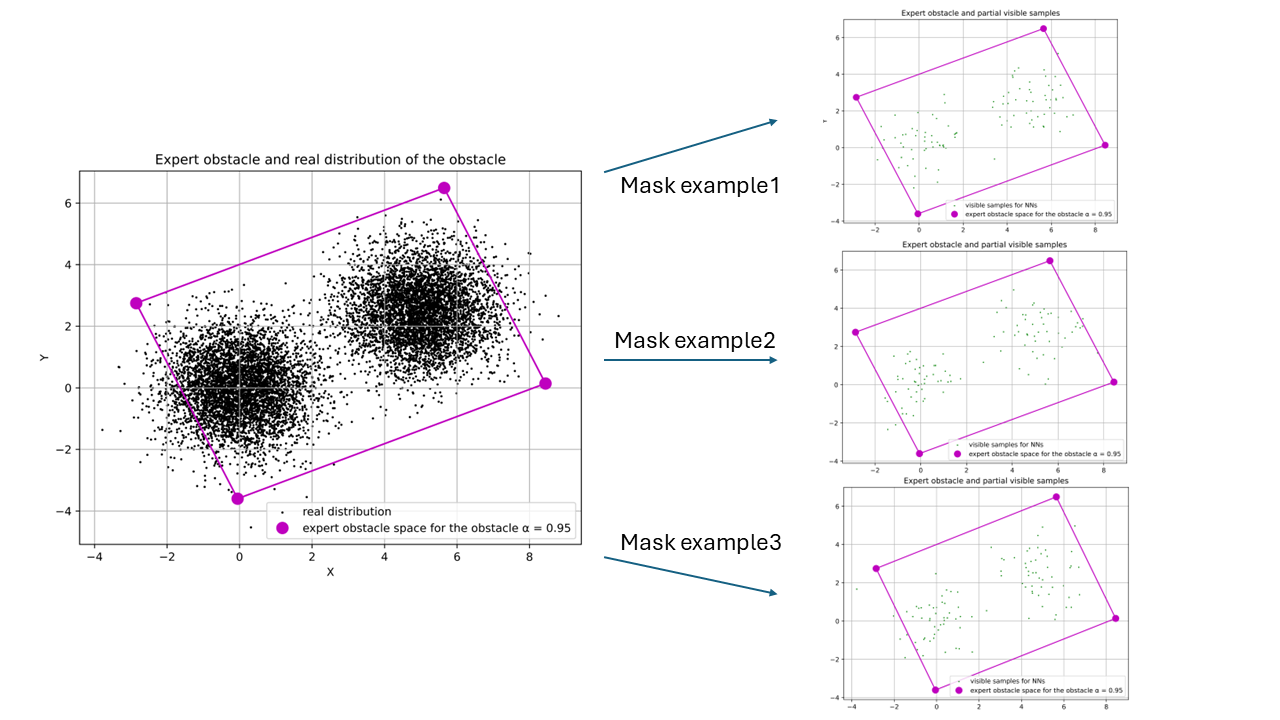}
    \caption{Supervised learning: we iteratively pick a few samples from $o^{true}$ as inputs to the neural network and task the network with generating the corresponding probabilistic obstacle space.}
    \label{Fig: maskedExample}
\end{figure*}
Previous research~\cite{he2022masked} has proven that the transformer can process the visible (unmasked) parts of the data, reconstructing the entire input from the encoded, incomplete data (masked data). Therefore, if we only have a few samples from a distribution, the transformer may be able to reconstruct the entire distribution. To train the neural network, we first require expert solutions generated by an expert method to serve as supervision during the training process.

In this paper, we employ the expert method, as proposed in our previous work~\cite{wu2023efficient}. The expert method can generate the probabilistic obstacle space of an uncertain object. This data-driven method builds on a kernel density estimator (KDE) that can generate the probabilistic obstacle space, accelerated by a fast Fourier transform (FFT). These KDE values are then utilized within an Integer Linear Programming (ILP) framework to find a minimal parallelogram box that approximates the probabilistic obstacle space. This ILP solution effectively encapsulates the area where the state is expected to stay with high confidence.

A brief process of the ILP Heuristic Algorithm is shown in the Algorithm~\ref{ag: ILP}.

\begin{algorithm}
\caption{ILP Heuristic Algorithm}\label{ag: ILP}
\begin{algorithmic}[1]
\Function{BoxApprox}{$r_{\min}, c_{\min}, r_{\max}, c_{\max}, \omega_{ij}$}
    \State $\omega'_{ij} \gets \omega_{ij}.\text{copy}()$
    \State $z_{ij}[1:N][1:N] \gets 0$
    \State $r_{\min}, c_{\min} \gets \arg \max_{i,j} \omega'_{ij}$
    \State $r_{\max} \gets r_{\min}$, $c_{\max} \gets c_{\min}$
    \While{$\sum_{i=1}^{N} \sum_{j=1}^{N} \omega_{ij} z_{ij} < \alpha \sum_{i} \sum_{j} \omega_{ij}$}
        \State $i, j \gets \arg \max_{i,j} \omega'_{ij}$
        \State $\omega'_{ij}[i, j] \gets 0.0$
        \State $r_{\min} \gets \min(r_{\min}, i)$, $r_{\max} \gets \max(r_{\max}, i)$
        \State $c_{\min} \gets \min(c_{\min}, j)$, $c_{\max} \gets \max(c_{\max}, j)$
        \State $z_{ij}[r_{\min}:r_{\max}][c_{\min}:c_{\max}] \gets 1$
    \EndWhile
    \State \Return $z_{ij}$
\EndFunction
\end{algorithmic}
\end{algorithm}

The ILP Heuristic Algorithm approximates solutions for integer linear programming problems using heuristics. It starts with a weight matrix $\omega_{ij}$ and a zero matrix $z_{ij}$, setting initial indices based on $\omega_{ij}$'s maximum values. The algorithm loops to update $z_{ij}$ until the product of $z_{ij}$ and $\omega_{ij}$ meets a threshold $\alpha$. Each iteration zeros out the maximum $\omega_{ij}$ and adjusts boundary indices. The corresponding $z_{ij}$ entry is set to one within these boundaries, yielding the matrix as an efficient approximation of the ILP solution. 

After obtaining the expert polygon vertices $\tilde{\mathcal{V}}$, we prepare the input data for network training. Fig.~\ref{Fig: maskedExample} illustrates how different subsets of samples are presented to the neural network during training. At each training step, a small random subset of samples is selected, and the transformer is trained to generate the corresponding polygon vertices. Through this supervised learning process, the network is repeatedly exposed to multiple partial views of the same obstacle, learning to associate incomplete sample sets with the full probabilistic obstacle space. At inference time, when only a limited number of samples are available, the trained network can infer the underlying distribution of the obstacle space by leveraging the patterns it has learned from these partial-to-full mappings.

\subsection{Multi-label learning and loss function}
\begin{figure}
    \centering
    \includegraphics[width=\linewidth]{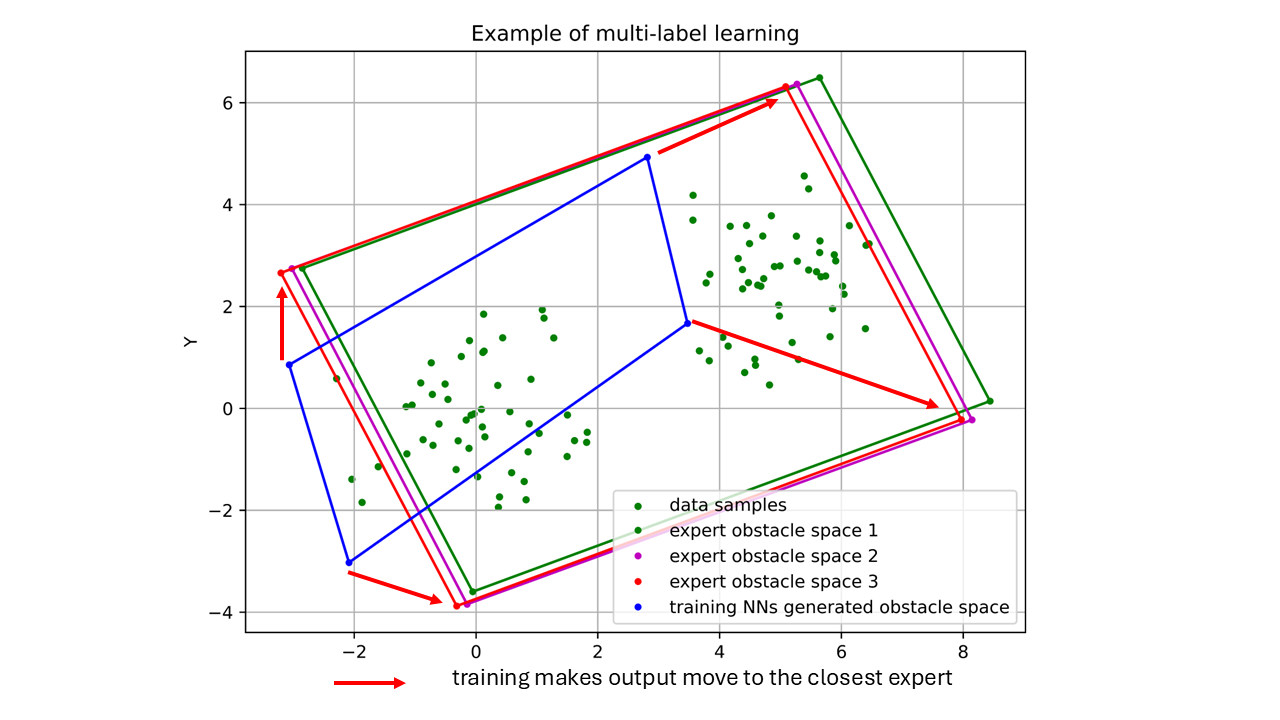}
    \caption{Multi-label learning: during training, the network is guided to produce outputs that align with the closest expert solution.}
    \label{Fig: multi_label_learning}
\end{figure}
Recall the non-uniqueness of confidence regions~\cite{neyman1937outline}: for any given distribution, there exist infinitely many subsets that capture the same probability mass. Consequently, for each uncertain obstacle, there are infinitely many valid probabilistic obstacle spaces. This motivates the use of multi-label learning rather than single-label learning. Multi-label learning not only reduces computational overhead but also allows the model to share learned features across multiple labels, potentially leading to better generalization. 
Fig.~\ref{Fig: multi_label_learning} illustrates this process. Given the current network parameters, the neural network produces an output polygon. The loss is then computed between this output and each available expert probabilistic obstacle space. The smallest loss among them is used to update the network weights. In practice, multiple expert probabilistic obstacle spaces are provided for each obstacle during training. Similar to many regression-based approaches, we employ the mean squared error (MSE) loss, as shown in Eq.~\ref{eq:loss}, which minimizes the average discrepancy between the network output and the expert labels.
\begin{equation}
\text{Loss} = \min_{j \in [1, 10]}{\frac{1}{8} \sum_{i=1}^{8} (\hat{y}_{ij} - y_{i})^2}
\quad \text{where} \quad 
\hat{y}_{ij} \in \mathbf{\tilde{\mathcal{V}}}
\label{eq:loss}
\end{equation}

where:
\begin{itemize}
    \item we have 8 (four 2d vertices) elements in the output vector,
    \item \(y_i\) is is the generated value for the \(i\)-th element,
    \item \(\hat{y}_{ij}\) is the \(j\)-th expert vertices for the \(i\)-th element.
\end{itemize}

\subsection{Neural networks architecture}
\begin{figure*}
    \centering
    \includegraphics[width=\linewidth]{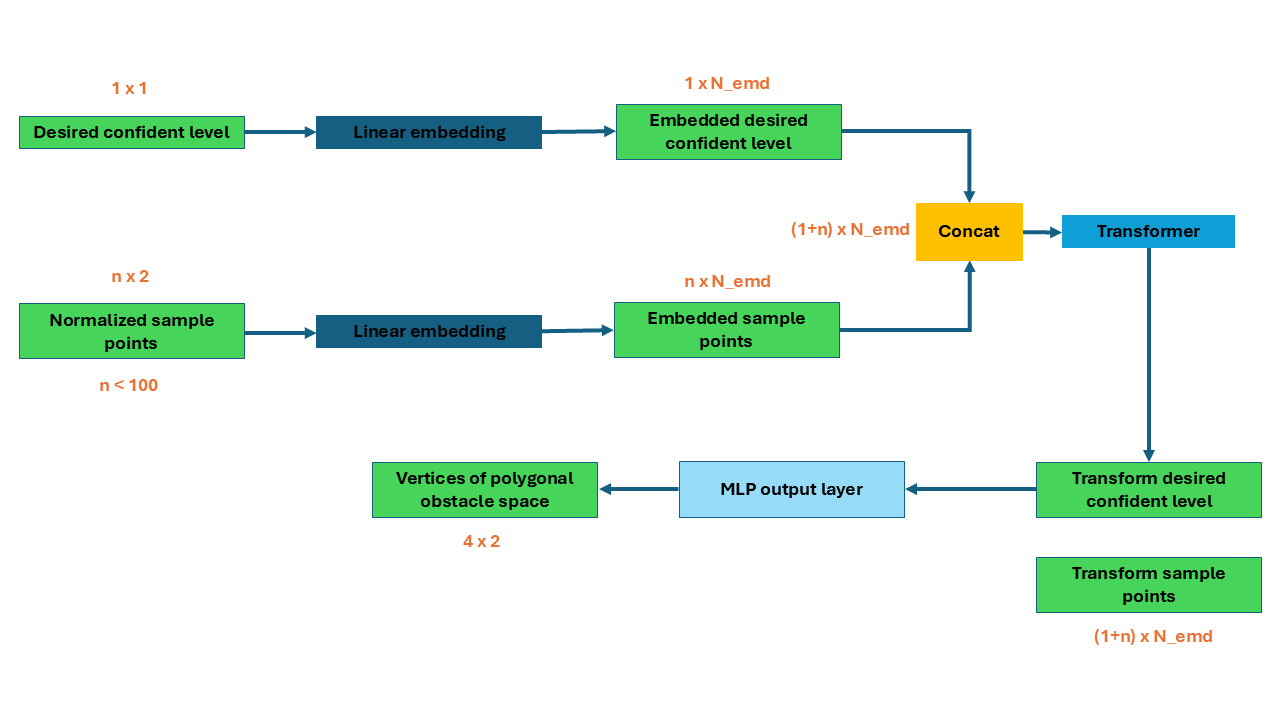}
    \caption{Neural network architecture}
    \label{Fig: NN architecture}
\end{figure*}

The overall architecture of the neural network we used is shown in Fig.~\ref{Fig: NN architecture}, which is designed based on our previous work~\cite{xiang2024imitation}. The input to the neural network consists of sampled points and the desired confidence level. Before sending these inputs to the neural network, we normalize them first. The confidence level and the sample points are then embedded in a higher dimension using a linear layer. The embedded items are concatenated together. These concatenated embeddings are then sent to the transformer layer. In the transformer, the channel representing the confidence level interacts with all the point channels. Finally, we send the confidence level channel to an output layer, resulting in four vertices of the convex polygon.

\begin{figure}
    \centering
    \includegraphics[width=\linewidth]{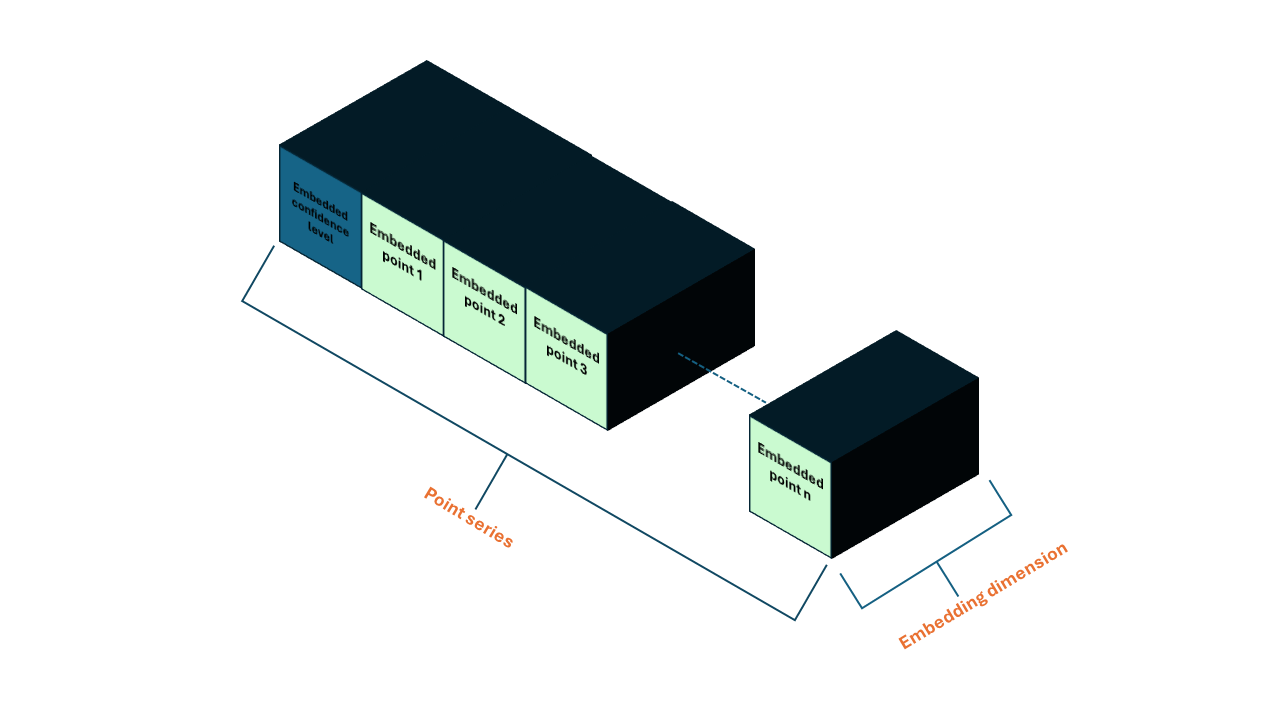}
    \caption{Concatenated tensor}
    \label{Fig: concatenate}
\end{figure}
\subsection{Input normalization, embedding, and concatenate}
Because sample points are drawn from different tasks, they have very different means and variances, which can cause challenges for the training process of neural networks. Normalizing those points before feeding them into the neural network is very crucial for several reasons. First, if some inputs have larger values and higher variance than others, they may dominate the training process. For example, if one target (or feature) has a very large value compared to others, its gradient will be disproportionately large. Second, normalization can accelerate the training process. When the scales of the inputs are not consistent, the model might spend more time adjusting weights to accommodate the varying scales of input data rather than learning meaningful patterns. Third, normalization can prevent overfitting by reducing bias toward dominant features and indirect regularization effects. 

We use the Min-Max scaling~\cite{hastie2009elements} to normalize the sample points. The Min-Max scaling equation is defined by: 
\begin{equation}
x' = \frac{x - \min(x)}{\max(x) - \min(x)}.
\end{equation}
Min-max scaling scales sample points to the range [0, 1] without changing the relative position between points. At the same time, since all the points are sampled from a reasonable distribution, so no outlier can heavily influence the normalization. 

After normalizing, we use a single linear layer to embed the inputs. Compared to vision or natural language inputs, the inputs in our task are much more straightforward to understand. We primarily aim to determine the relationship between points rather than to interpret their meanings. Additionally, the transformer model already incorporates non-linearities. Therefore, we choose to embed the inputs into relatively low dimensions and use only a single linear layer to decrease the computing complexity. 

After embedding, the point tensor has dimensions ($N$, $C$), where $N$ is the dimension of the series of points and $C$ is commonly referred to as the channel in the field of machine learning. We concatenated the confidence level tenor and the point tensor together along the $N$ dimension. So, the concatenated tensor has dimensions ($N+1$, $C$), while the first channel is the channel of the confidence level. The concatenated tensor then looks like the Fig.~\ref{Fig: concatenate}.

\subsection{Transformer}
For this work, we only use the transformer encoder~\cite{vaswani2017attention} because our task requires predicting all polygon vertices at once, instead of generating them sequentially.  The encoder is composed of a stack of 6 identical layers, each layer has two sub-layers, a multi-head self-attention mechanism, and a feedforward network. We also kept the residual connection and layer normalization.
Compared to other popular layers, such as the recurrent and convolutional layers, self-attention has three main advantages. 
First, self-attention has lower computational complexity per layer. Second, it requires fewer sequential operations since most computations can be parallelized. The third advantage is that self-attention can learn the dependencies between two points that are far from each other. For example, if we use CNNs, we may input points patch by patch, then it is difficult to learn the relationship between the first patch and the last patch. In contrast, each point interacts with all the other points at the same time in self-attention. 

\subsection{Output}


In prior masked autoencoder research, mask tokens are introduced after the encoder so that the decoder can identify which missing positions need to be reconstructed. Our setting is fundamentally different. Unlike reconstruction tasks, we do not use mask tokens, since the visible samples do not carry positional information about which part of the obstacle space they come from. Moreover, sequence order is irrelevant in our problem. In natural language processing (NLP), for example, the sentences ‘The dog chased the cat’ and ‘The cat chased the dog’ convey completely different meanings despite containing the same words. In contrast, for our problem, the point sets ${(1,1),(2,2)}$ and ${(2,2),(1,1)}$ represent the same distribution. Consequently, no additional global context is required. Furthermore, the objective of our neural network is not to reconstruct the full underlying distribution but to generate a convex bounding region. Thus, a simple output layer suffices to produce the convex bound from the features extracted by the encoder.

We use a multilayered output like ViT~\cite{dosovitskiy2020image} instead of a single linear layer like the original transformer. The multi-layer output can further process the features extracted from the point channels and refine the final classification. We only used the transformed confidence level channel for output. The transformer allows the confidence level channel to attend to all other channels (embedded points), gathering contextual information from all the points.  
Finally, we need to denormalize the output, since the points we want to bound are normalized.

\subsection{Parameters}
For the transformer, the embedding dimension $N_{emd}$ is 864. We use 6 attention heads, so each head processes 144 embedding dimensions of each channel. The number of attention layer is also 6. Optimization is performed using the Adam optimizer with a dynamic learning rate to ensure gradual convergence. Additionally, a dropout rate of 0.2 is applied to prevent overfitting by randomly omitting certain neurons during training. 

\section{Result}

\begin{figure*}[!t]
    \centering
    \subfloat[motion uncertainty\label{fig:motion}]{%
        \includegraphics[width=0.29\textwidth]{figure/motion_result_example0.95.png}%
    }\hfill
        \subfloat[behavior uncertainty\label{fig:behavior}]{%
        \includegraphics[width=0.29\textwidth]{figure/behavior_result_example0.99.png}%
        \label{fig:sub3}
    }\hfill
    \subfloat[sensor uncertainty\label{fig:sensor}]{%
        \includegraphics[width=0.29\textwidth]{figure/detection_result_example0.8.png}%
    }
    \caption{Black points are sampled from the real obstacle distribution. Colorful polygons are example outputs when the three methods take the same yellow visible points. Yellow points are enlarged for visibility.}
    \label{fig:generating_demo}
\end{figure*}

\subsection{Experiment setting}
To evaluate the effectiveness of the proposed method, we conduct experiments on three types of uncertain obstacles: flying obstacles subject to motion uncertainty, obstacles with uncertain destinations, and obstacles whose positions are estimated using noisy sensor measurements. 
We feed the neural network with samples drawn from the true obstacle distribution and desired confidence level $\alpha$, then evaluate whether the generated probabilistic obstacle space achieves the desired coverage, i.e., whether it bounds the specified proportion of the distribution. Obstacles of the same type share the same underlying distribution but may exhibit different geometric shapes.

\subsubsection{Training}
We trained a neural network that can work with varying values of $\alpha$, where $\alpha$ can be 80\%, 85\%, 90\%, 95\% and 99\%. To accelerate the training process, we prepare 100 sets of expert vertices for each desired confidence level and prepare 320,000 sets of 100 samples before training began. In each training iteration, we randomly select one set of samples and one of $\alpha$ as the neural network's input, then randomly assign ten sets of corresponding expert vertices as the target.  We can prepare the training dataset for a known uncertain obstacle within a few hours. It supports the \textit{Advantage 3} of the proposed method, the training dataset is easy to make.

\subsubsection{Interface}
We tested the trained neural networks by presenting them with 5,000 random tasks (1000 for each $\alpha$) from each type of uncertain obstacle. Each task contains 100 samples randomly generated from the real distribution of the uncertain obstacle and has never been seen by the neural network. The neural network generates the corresponding $o^{\alpha}$ for those samples. 

\subsubsection{evaluating}
To evaluate the $o^{\alpha}$ generated by the neural network, we first need to determine the coverage. 
We approximate the true obstacle distribution using 100,000 samples, as the underlying distributions are typically empirical. The resulting coverage was computed as the percentage of samples that fell within the generated $o^{\alpha}$.
We use mean squared error (MSE) to quantify the accuracy of the $o^{\alpha}$:
\begin{equation}
\text{MSE} = \frac{1}{n} \sum_{i=1}^n \left( \alpha_i - \hat{\alpha}_i \right)^2.
\end{equation}
$\alpha_i$ is the real coverage of the generated $o^{\alpha}$ and $\hat{\alpha}_i$ is the desired confidence level (CL). n is 1000 because we test 1000 times for each $\alpha$. We also recorded the minimum and maximum coverage values to evaluate performance stability. While a low MSE indicates that the probabilistic obstacle space is accurate, it does not guarantee correctness. Therefore, it is essential to compare the percentage of instances where $\alpha_i$ exceeds $\hat{\alpha}_i$. We called it the correctness rate:
\begin{equation}
    \text{Correctness Rate} = \frac{1}{n} \sum_{i=1}^{n} \mathbf{1}\!\left[ \alpha_i > \hat{\alpha}_i \right],
\end{equation}
where $1[\bullet]$ is the indicator function(1 if the condition is true, 0 otherwise).
\subsubsection{baseline}
We compare the proposed method with the ILP Heuristic (ILP), which is the expert method, and the bounding box (BB) method~\cite{zhao2012algorithm}\cite{ericson2004real}, a popular approach to finding a polygonal approximation of the obstacle space. Both baseline methods are KDE-based. We generate $o^{\alpha}$ with BB and ILP for the same obstacle as the baseline. Fig.~\ref{fig:generating_demo} are examples of obstacle distribution and $o^{\alpha}$ generated by different methods given the same samples. However, since BB and ILP are not learning-based methods and do not require a training process, a direct comparison may not be entirely fair. To address this, we also provide the baseline methods with 20 times more samples to enable a more thorough evaluation of our approach. In addition to these baselines, we implement the MINLP Optimal Algorithm~\cite{wu2024data2} to compute the smallest convex region that satisfies the coverage requirement.  All the experiences are run in Intel(R) Core(TM) i9-12900KF and NVIDIA GeForce RTX 3090.

\begin{table*}
    \centering
    \small
    \caption{MSE and range on motion uncertainty distribution compared to baseline with a small number of samples}
    \label{tab: accuracymotionsmall}
        \begin{tabular}{|c|c|c|c|}
    \hline 
        & \multicolumn{1}{c|}{NN(our method)} & \multicolumn{1}{c|}{BB(100)} & \multicolumn{1}{c|}{ILP(100)} \\
          \hline 
        CL & MSE(range) &MSE(range)&MSE(range)\\ 
     \hline 
      80\% &  0.32\%(78.95\% - 80.44\%)&11.98\%(80.28\% - 99.80\%)&2.83\%(75.04\% - 91.87\%)\\ 
      85\% &  0.25\%(84.44\% - 86.21\%)&9.88\%(85.39\% - 99.91\%)&
      2.95\%(75.24\% - 94.01\%) \\ 
      90\% &  0.23\%(89.40\% - 90.91\%)&7.36\%(89.60\% - 99.99\%)&
      2.69\%(75.63\% - 98.43\%)\\ 
      95\% &  0.49\%(95.22\% - 96.02\%)&4.32\%(92.60\% - 100\%)&
      2.20\%(85.67\% - 99.88\%) \\ 
      99\% &  0.98\%(99.96\% - 99.99\%)& 1.33\%(98.03\% - 100\%)&
       1.12\%(97.86\% - 100\%)\\ 
     \hline
    \end{tabular}
    \vspace{1em}
        \small
        \caption{MSE and range on motion uncertainty compare to baseline with a large number of sample}
        \label{tab: accuracymotionlarge}
    \begin{tabular}{|c|c|c|c|}
    \hline 
        & \multicolumn{1}{c|}{NN(our method)}  & \multicolumn{1}{c|}{BB(2000)}& \multicolumn{1}{c|}{ILP(2000)} \\
          \hline 
        CL & MSE(range) &MSE(range)&MSE(range) \\ 
     \hline 
      80\% &  0.32\%(78.95\% - 80.44\%)&12.73\%(88.66\% - 96.29\%)&2.01\%(75.36\% - 84.46\%) \\ 
      85\% &  0.25\%(84.44\% - 86.21\%)&10.98\%(91.93\% - 98.82\%)&
      2.31\%(81.43\% - 90.18\%) \\ 
      90\% &  0.23\%(89.40\% - 90.91\%)&8.63\%(95.93\% - 99.75\%)&
      1.90\%(87.40\% - 94.16\% )\\ 
      95\% &  0.49\%(95.22\% - 96.02\%)&4.81\%(98.59\% - 100\%)&2.72\%(94.35\% - 99.06\%) \\ 
      99\% &  0.98\%(99.96\% - 99.99\%)&0.99\%(99.96\% - 100\%)&
       0.98\%(99.19\% - 100\%)\\ 
     \hline
    \end{tabular}
    \vspace{1em}
        \caption{correctness rate/Probabilistic obstacle space area for each desired confidence level on motion uncertainty}
    \label{tab:correctmotion}
\begin{tabular}{|c|c|c|c|c|c|c|}
\hline 
   & \multicolumn{6}{c|}{Correctness rate}  \\ 
 \hline
CL& \textbf{NN}&BB(100)&BB(2000)&ILP(100)&ILP(2000)&optimal(10000)\\
\hline  
 80\% & 98.7\%&100\%&100\%&57.7\%&79.8\%&100\%  \\ 
 85\% & 99.8\%&100\%&100\%&52.4\%&64.4\%&100\%\\
 90\% & 99.7\%&99.8\%&100\%&52.8\%&82.3\%& 100\% \\
 95\% & 100\%&99.4\%&100\%&74.8\%&98.9\%& 100\% \\
 99\% & 100\%&61\%&100\%&59.7\%&100\%& 100\%\\
\hline 
 CL  & \multicolumn{6}{c|}{Probabilistic obstacle space area}  \\ 
 \hline
 80\% &1415.07&2342.52&2309.52&1783.52&1514.04&1172.71\\ 
 85\% &1762.30&2790.63&2815.55&2138.71&1902.27&1579.52\\
 90\% &2229.01&3367.96&3470.37&2652.60&2365.27&2107.79 \\
 95\% &3530.87&4525.46&4576.97&3382.33&3374.61&2742.49 \\
 99\% &4833.31&5749.98&6127.58&4967.29&4820.67&3711.22\\
\hline

\end{tabular}

\end{table*}

\begin{table*}
\small
\centering
    \caption{MSE and range on behavior uncertainty compare to baseline with a small number of sample}
    \label{tab: accuracyGMRsmall}
\begin{tabular}{|c|c|c|c|c|c|}
        \hline 
       & \multicolumn{1}{c|}{NN(our method)} & \multicolumn{1}{c|}{BB(300)} & \multicolumn{1}{c|}{ILP(300)}\\ 
          \hline 
        CL & MSE(range) &MSE(range)&MSE(range)\\ 
     \hline 
      80\% & 1.79\%(79.54\% - 84.27\%) & 11.31\%(84.04\% - 96.41\%)&2.74\%(75.00\% - 88.70\%)\\ 
      85\% & 1.76\%(82.34\% - 88.25\%) & 9.56\%(88.09\% - 98.02\%)&2.62\%(75.31\% - 90.94\%) \\ 
      90\% & 1.13\%(88.31\% - 91.8\%)& 7.37\%(93.11\% - 99.40\%)&1.76\%(81.48\% - 96.56\%) \\ 
      95\% &  0.85\%(94.14\$ - 97.83\%) & 4.09\%(95.82\% - 99.95\%)&1.11\%(90.61\% - 98.80\%) \\ 
      99\% &  0.58\%(99.18\% - 99.94\%) & 0.80\%(98.42\% - 100\%) &0.40\%(97.03\% - 99.95\%)  \\ 
     \hline
    \end{tabular}
    \vspace{1em}
    \small
        \caption{MSE and range on behavior uncertainty compare to baseline with a large number of sample}
    \label{tab: accuracyGMRlarge}
    \begin{tabular}{|c|c|c|c|}
        \hline 
       & \multicolumn{1}{c|}{NN(our method)} &  \multicolumn{1}{c|}{BB(6000)}&  \multicolumn{1}{c|}{ILP(6000)}\\ 
          \hline 
        CL & MSE(range) &MSE(range)&MSE(range)\\ 
     \hline 
      80\% & 1.79\%(79.54\% - 84.27\%) &10.47\%(85.60\% - 93.33\%)&2.40\%(75.00\% - 81.90\%)\\ 
      85\% & 1.76\%(82.34\% - 88.25\%) &9.45\%(90.69\% - 96.35\%)&3.01\%(85.60\% - 93.33\%) \\ 
      90\% & 1.13\%(88.31\% - 91.8\%)& 7.44\%(95.89\% - 98.67\%)&1.37\%(85.62\% - 94.55\%) \\ 
      95\% & 0.85\%(94.14\$ - 97.83\%) & 4.22\%(97.98\% - 99.67\%)&0.74\%(92.35\% - 97.50\%) \\ 
      99\% & 0.58\%(99.18\% - 99.94\%) & 0.91\%(99.72\% - 99.97\%) & 0.25\%(98.28\% - 99.61\%) \\ 
     \hline
     
    \end{tabular}
    \vspace{1em}
\small
    \caption{correctness rate/Probabilistic obstacle space area for each desired confidence level on behavior uncertainty}
    \label{tab:correct3}
\begin{tabular}{|c|c|c|c|c|c|c|}
\hline 
   & \multicolumn{6}{c|}{Correctness rate}  \\ 
 \hline
CL& \textbf{NN}&BB(300)&BB(6000)&ILP(300)&ILP(6000)&optimal(30000)\\
\hline  
 80\% & 93.1\%&100\%&100\%&10.8\%&3.9\%&100\%  \\ 
 85\% & 95.2\%&100\%&100\%&27.9\%&16.2\%& 100\%\\
 90\% & 99.8\%&100\%&100\%&69.2\%&55.8\%& 100\%\\
 95\% & 95.6\%&100\%&100\%&68.2\%&65.4\%& 100\% \\
 99\% & 100\%&99.7\%&100\%&56.5\%&75.8\%& 100\%\\
\hline 
 CL  & \multicolumn{6}{c|}{Probabilistic obstacle space area}  \\ 
 \hline
 80\% &2.75E-3&3.64E-3&3.46E-3&2.59E-3&2.39E-3&2.15E-3  \\ 
 85\% &3.44E-3&4.41E-3&4.35E-3&2.99E3&2.89E-3&2.39E-3\\
 90\% &4.41E-3&7.37E-3&5.81E-3&3.83E-3&3.67E-3&2.99E-3 \\
 95\% &5.69E-3&7.89E-3&7.98E-3&5.03E-3&4.87E-3& 3.95E-3 \\
 99\% &9.21E-3&11.19E-3&11.16E-3&8.01E-3&7.63E-3& 6.46E-3\\
\hline

\end{tabular}

\end{table*}

\begin{table*}
    \centering
        \small
    \caption{MSE and range on sensor uncertainty distribution compare to baseline with a small number of sample}
    \label{tab: accuracysensorsmall}
    \begin{tabular}{|c|c|c|c|}
    \hline 
       & \multicolumn{1}{c|}{NN(our method)} & \multicolumn{1}{c|}{BB(100)} & \multicolumn{1}{c|}{ILP(100)} \\
     \hline 
        CL & MSE(range) &MSE(range)&MSE(range) \\ 
     \hline 
      80\% & 0.42\%(79.12\% - 82.05\%) &11.04\%(87.56\% - 99.42\%) &3.66\%(72.00\% - 93.54\%) \\ 
      85\% & 0.35\%(84.45\% - 86.63\%)&10.18\%(91.98\% - 99.80\%)&3.70\%(72.17\% - 95.76\%) \\ 
      90\% & 0.24\%(89.51\% - 91.01\%)&8.20\%(93.26\% - 99.94\%)&4.98\%(86.16\% - 98.78\%) \\ 
      95\% & 0.62\%(94.08\% - 96.29\%)&4.20\%(94.84\% - 99.99\%)&3.25\%(92.45\% - 99.73\%)\\ 
      99\% & 0.87\%(97.66\% - 99.94\%)&1.87\%(97.09\% - 99.99\%)&0.80\%(98.97\% - 99.99\%) \\ 
    \hline
    \end{tabular}
    \vspace{1em}
    
    \small
    \caption{MSE and range on sensor uncertainty distribution compare to baseline with a large number of sample}
    \label{tab: accuracysensorlarge}
    \begin{tabular}{|c|c|c|c|}
    \hline 
       & \multicolumn{1}{c|}{NN(our method)} & \multicolumn{1}{c|}{BB(2000)}& \multicolumn{1}{c|}{ILP(2000)} \\
     \hline 
        CL & MSE(range) &MSE(range)&MSE(range) \\ 
     \hline 
      80\% & 0.42\%(79.12\% - 82.05\%) &16.26\%(93.24\% - 98.43\%)&4.12\%(72.01\% - 86.43\%) \\ 
      85\% & 0.35\%(84.45\% - 86.63\%)&12.10\%(95.65\% - 98.49\%)&3.15\%(74.27\% - 93.60\%) \\ 
      90\% & 0.24\%(89.51\% - 91.01\%)&8.12\%(96.52\% - 99.63\%)&4.37\%(89.70\% - 98.32\%) \\ 
      95\% & 0.62\%(94.08\% - 96.29\%)&4.28\%(98.64\% - 99.86\%)&3.22\%(95.48\% - 99.53\%) \\ 
      99\% & 0.87\%(97.66\% - 99.94\%)&0.93\%(99.73\% - 99.99\%)&0.86\%(99.49\% - 99.98\%) \\ 
    \hline
    \end{tabular}
    \vspace{1em}

    \centering
\small

\caption{correctness rate/Probabilistic obstacle space area for each desired confidence level on sensor uncertainty}
    \label{tab: correctsensor}
        \centering
    \small
    \begin{tabular}{|c|c|c|c|c|c|c|}
\hline 
   & \multicolumn{6}{c|}{Correctness rate}  \\ 
 \hline
CL& \textbf{NN}&BB(100)&BB(2000)&ILP(100)&ILP(2000)&optimal(10000)\\
\hline  
 80\% & 99.7\%&100\%&100\%&31.4\%&14.7\%&100\%  \\ 
 85\% & 99.8\%&100\%&100\%&67.3\%&57.9\%&100\%\\
 90\% & 99.8\%&100\%&100\%&97.2\%&99.8\%& 100\% \\
 95\% & 98.9\%&99.9\%&100\%&99.1\%&100\%& 100\% \\
 99\% & 98.7\%&95.2&100\%&99.9\%&100\%& 100\%\\
\hline 
 CL  & \multicolumn{6}{c|}{Probabilistic obstacle space area}  \\ 
 \hline
 80\% & 35.22&45.65&46.14&36.06&33.95&33.01  \\ 
 85\% & 41.58&52.16&50.12&42.42&40.69& 34.23\\
 90\% & 51.38&63.18&59.10&53.86&52.35& 42.79 \\
 95\% & 64.53&75.85&71.90&67.46&66.08& 55.01 \\
 99\% & 99.03&101.49&106.67&95.28&96.10& 73.96\\
\hline
\end{tabular}

\end{table*}

\subsection{Motion Uncertainty}
For motion uncertainty, we evaluate flying obstacles subject to stochastic deviations in their trajectories. We simulate a flight using the Dubins vehicle model, where both speed and heading angle are sampled from a truncated Gaussian distribution. The resulting obstacle distribution for this simulated flight is shown in Fig.~\ref{fig:motion}.

From Tables~\ref{tab: accuracymotionsmall} and~\ref{tab: accuracymotionlarge}, our method demonstrates remarkable accuracy even with a small number of samples. For instance, at 90\% confidence level (CL), NN achieves an MSE of 0.23\% with a range of 89.40\%--90.91\%, whereas BB with 100 samples exhibits a much larger error of 7.36\% and wider range (89.60\%--99.99\%). Even when the baselines are provided with 20$\times$ more samples, NN maintains superior accuracy and stability.  

Correctness rates in Table~\ref{tab:correctmotion} further highlight the robustness of our approach. NN achieves above 98\% correctness at all CLs, while ILP often falls below 60\% with limited samples. BB can achieve high correctness, but only at the cost of significantly larger probabilistic obstacle space areas, which reduces path planning efficiency. In contrast, NN produces obstacle spaces closer to the optimal size, especially at higher CLs.

\subsection{Behavior Uncertainty}
For behavior uncertainty, we evaluate flying obstacles with uncertain destinations. The real obstacle distribution is generated using a Gaussian Mixture Regression (GMR) trajectory predictor, which estimates the probability distribution of a flight’s future positions in terminal airspace~\cite{xiang2024data}. Given the past trajectory, the predictor models multiple possible future routes as a mixture of Gaussians. The resulting obstacle distribution for this predicted flight is shown in Fig.~\ref{fig:behavior}.

Results under behavior uncertainty (Tables~\ref{tab: accuracyGMRsmall}--\ref{tab:correct3}) reveal the strength of the learning-based approach. Despite the increased distributional complexity, NN maintains low MSE (below 2\% for CL $\geq$ 85\%) and narrow coverage ranges. By comparison, BB consistently overshoots the target coverage, while ILP suffers from high variance and instability, particularly with limited data.  

The correctness rates again demonstrate the advantage of NN. At 90\% CL, NN achieves nearly perfect correctness (99.8\%), whereas ILP with 300 samples drops to 69.2\%. Importantly, NN generates obstacle spaces that are consistently smaller than BB while maintaining high coverage reliability. 

\subsection{Sensor Uncertainty}
For sensor uncertainty, we evaluate obstacles whose positions are estimated from noisy measurements. In this scenario, the true positions of two adjacent obstacles are observed through a sensor corrupted by Gaussian noise, resulting in uncertainty about their exact locations. The real obstacle distribution is obtained by sampling noisy measurements of the true positions and constructing the corresponding probabilistic obstacle space. The resulting distribution is illustrated in Fig.~\ref{fig:sensor}.

Under sensor uncertainty (Tables~\ref{tab: accuracysensorsmall}--\ref{tab: correctsensor}), NN achieves the lowest MSE across all CLs, e.g., 0.24\% at 90\% CL, compared to 8.20\% for BB (100 samples) and 4.98\% for ILP (100 samples). Even with 20$\times$ more samples, BB and ILP fail to match NN’s accuracy and stability.  

Correctness rates remain above 98\% for NN, ensuring that the generated obstacle spaces are reliable. Moreover, the obstacle space areas are significantly closer to the optimal solution compared with BB and ILP. This result keeps demonstrating NN’s ability to generalize well under noisy measurement conditions while avoiding excessive conservatism.

\subsection{Overall Discussion}
Across motion, behavior, and sensor uncertainty, the proposed NN consistently outperforms traditional BB and ILP methods. Three key observations can be drawn:
\begin{itemize}
    \item \textbf{Accuracy with Few Samples:} NN achieves high accuracy and low MSE even with very limited data, whereas baselines require 20$\times$ more samples to approach similar performance.  
    \item \textbf{Reliability:} NN maintains correctness rates above 93\% in all scenarios, ensuring that the generated probabilistic obstacle spaces meet or exceed the desired confidence levels.  
    \item \textbf{Efficiency:} NN produces compact obstacle spaces that are significantly smaller than BB while avoiding the instability of ILP. This efficiency is particularly critical in path planning applications where over-approximation reduces safe space.  
\end{itemize}

These results validate the effectiveness of the proposed learning-based framework in generating accurate and reliable probabilistic obstacle spaces under diverse forms of uncertainty.

\section{Generating speed analysis}
\begin{figure}
    \includegraphics[width=\linewidth]{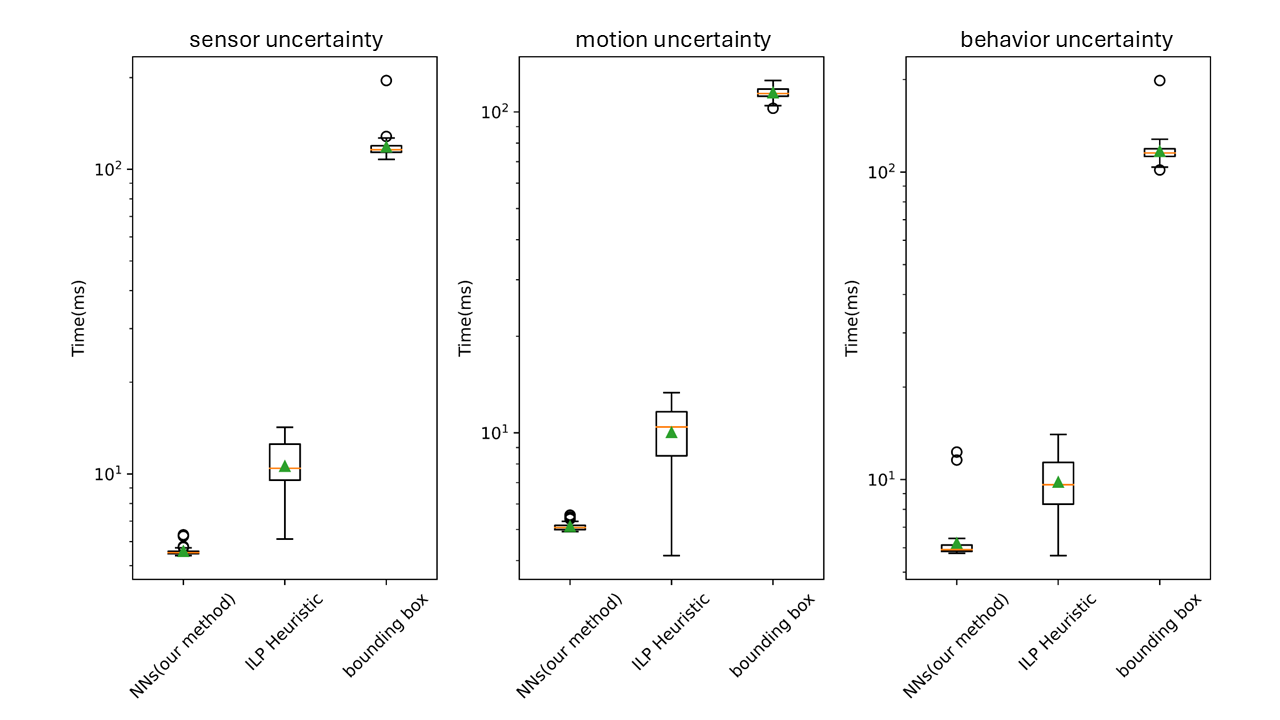}
    \caption{Generating speed comparison}
    \label{Fig: generating_spped}
\end{figure}

Generating time is crucial in deciding whether a method can be used in online applications. Therefore, we compare the time to generate one correct probabilistic obstacle space by each method to prove our method is more useful in online applications. Fig.~\ref{Fig: generating_spped} illustrates the average time cost for generating effective outcomes. The results clearly demonstrate that our method significantly outperforms the baseline approaches, achieving execution times that are an order of magnitude faster. The result supports that the \textit{Advantage 4} of the proposed method, NN can generate a useful probabilistic obstacle space very fast. 

\section{Conclusion and Future }

In this paper, we propose using supervised learning to train a transformer-based neural network to generate a confidence region of the real obstacle space distribution, which we define as the probabilistic obstacle space.  Once trained, the neural network can robustly generate vertices of the polygon that cover the desired percentage of the real obstacle region, only given a few samples. The primary advantage of the proposed method is its extremely fast generation time while maintaining performance comparable to that of the baseline methods. Meanwhile, training is also very simple because training inputs can be generated very fast. Only a few experts are required for the labels. No dynamic information about the obstacle such as disturbance is required.

\section*{Acknowledgment}
\medskip

The authors would like to acknowledge the support from the National Science Foundation under Grants CCF-2402689. Any opinions, findings, conclusions, or recommendations expressed in this paper are those of the authors and do not reflect the views of NSF.


\bibliography{main}
\bibliographystyle{IEEEtran}


\section{Biography Section}
\begin{IEEEbiography}[{\includegraphics[width=1in,height=1.25in,clip,keepaspectratio]{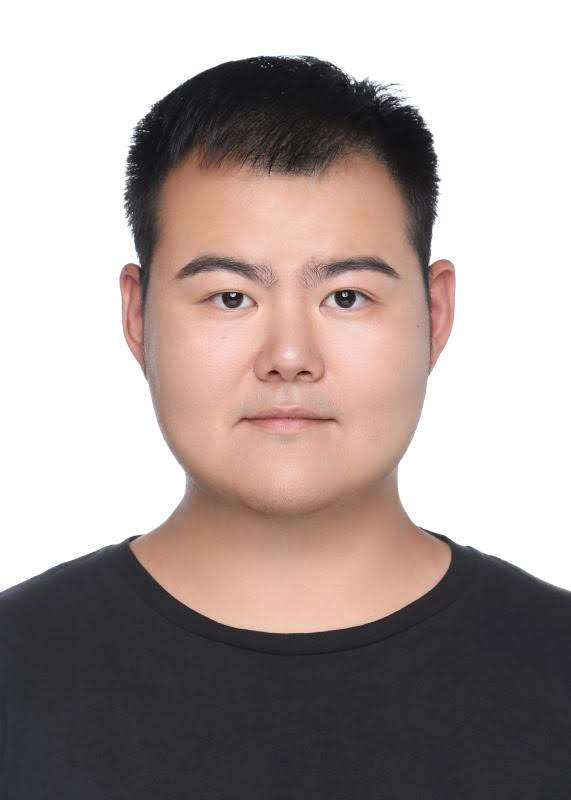}}]{Jun Xiang} is currently pursuing the joint
Ph.D. degree at the Department of Mechanical
and Aerospace Engineering, University of California
at San Diego, and the Department of Aerospace
Engineering, San Diego State University. He received his M.S. degree in Automotive Engineering from Clemson University and his B.S. degree in Mechanical Engineering from North Carolina State University. His research interests include motion planning and prediction of autonomous air/ground vehicle systems.\end{IEEEbiography}

\begin{IEEEbiography}[{\includegraphics[width=1in,height=1.25in,clip,keepaspectratio]{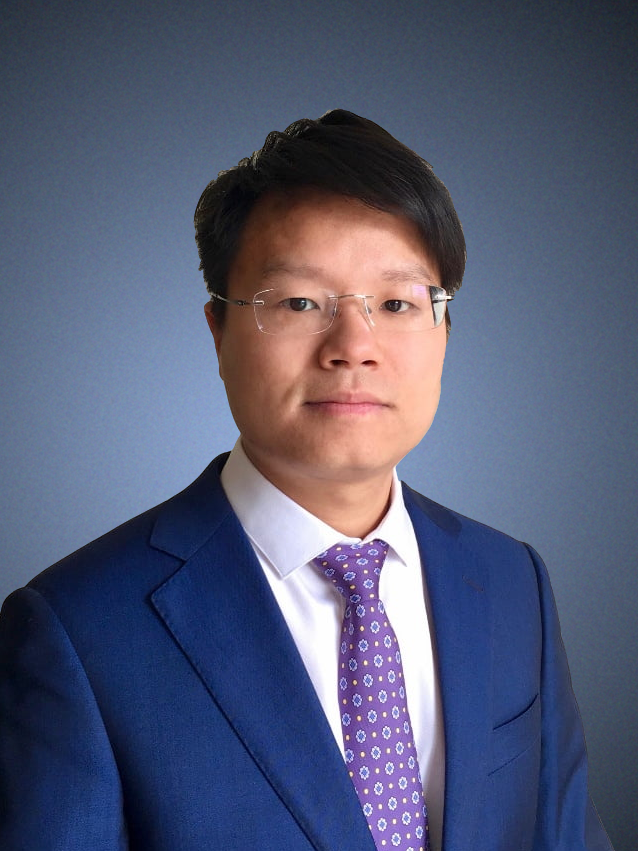}}]{Jun Chen}is an Associate Professor of Aerospace Engineering at San Diego State University. He received his M.S. and Ph.D. degrees in Aerospace Engineering from Purdue University. His research interests include control and optimization for large-scale networked dynamical systems, with applications in mechanical and aerospace engineering such as air traffic control, traffic flow management, and autonomous air/ground vehicle systems. 
\end{IEEEbiography}

\vfill

\end{document}